# A Hybrid Precipitation Prediction Method based on Multicellular Gene Expression Programming

Li Hongya[1], PENG Yuzhong[1,2+], Deng Chuyan[1], Pan Yonghua, Gong Daoqing[1], Zhang Hao

1. School of Computer & Information Engineering, Guangxi Teacher Education University, Nanning 530001, China
2. School of Computer science, Fudan University, Shanghai, 200433, China
3. Department of science, Shangqiu University Applied Science and Technology College, Kaifeng 475000, China
+ Corresponding author: E-mail: jedison@163.com

**Abstract**：Prompt and accurate precipitation forecast is very important for development management of regional water resource, flood disaster prevention and people's daily activity and production plan; however, non-linear and nonstationary characteristics of precipitation data and noise seriously affect forecast accuracy. This paper combines multicellular gene expression programming with more powerful function mining ability and wavelet analysis with more powerful denoising and extracting data fine feature capability for precipitation forecast modeling, proposing to estimate meteorological precipitation with WTGEPRP algorithm. Comparative result for simulation experiment with actual precipitation data in Zhengzhou, Nanning and Melbourne in Australia indicated that: fitting and forecasting performance of WTGEPRP algorithm is better than the algorithm Multicellular Gene Expression Programming-based Hybrid Model for Precipitation Prediction Coupled with EMD, Supporting Vector Regression, BP Neural Network, Multicellular Gene Expression Programming and Gene Expression Programming, and has good application prospect.

## 1 引言

降水是自然界中的重要气候现象，作为大气循环的重要组成部分，降水在多种复杂因素共同作用下形成。短时间大量集中降水，极易形成洪涝灾害，影响国计民生。因此，及时研究降水量的变化，精确预知未来降水情况，对做好日常生活和生产规划提供依据，对合理开发利用水资源和防洪抗旱提供可靠参考，减少不必要的损失，为国计民生的发展提供保障。

传统降水预测模型，如回归方程预测法[1]等在分析处理大量降雨时间序列数据时，不仅不易准确挖掘数据中的一些关键信息，还会引入噪声或者造成信息冗余而直接影响预测质量，预测精度较低[2]。随着气象、人工智能和数据挖掘研究的发展，运用智能计算和数据挖掘技术进行地区降水预测，为深入挖掘大量降水时间序列数据的内在规律提供了新的有效方法，成为研究热点问题之一。近年，可以对降水时间序列数据内部各影响要素之间的复杂关系进行有效描述的神经网络方法和SVM算法已被广泛应用在降水建模预测中。如：Devi S R[3]等对BPN、CBPN、DTDNN和NARX神经网络模型预测能力的比较，证明NARX模型的预测性能优良。孟锦根等[4]提出了PSO-LSSVM算法，用PSO寻找SVM的最优参数，并用新疆阿勒泰的降水数据证明算法确实有更好的预测精度；Shamshirband S[4]等提出ANFIS和SVR结合的降水预测方法。通过自塞尔维亚29个气象站的月降

水量数据进行模拟实验，实验结果充分证明了算法的有效性。总的来说，神经网络和 SVM 算法虽然可对降水时间序列数据内部各影响要素之间的复杂关系进行有效描述，但却很难统一选定算法本身的结构和参数，且算法本身的计算量过大对大容量样本的训练学习是不利的。

基因表达式编程（Gene Expression Programming，GEP）算法，是融合了遗传算法和遗传编程的个体组织方法的优点而提出的新的自适应演化算法。其对符号和表达式的处理优于遗传算法，性能上高于遗传编程，编码简单，可对任何的复杂问题进行编码，能在全局空间内进行搜索，寻优能力、发现规律和公式的能力强。已有的研究工作表明，GEP 算法具有非常强的回归分析、求解问题的能力，特别在传统方法很难处理的、需要全局优化的非线性复杂问题方面有较强的优势[6-8]。但是，目前用 GEP 相关算法直接对降雨量序列数据本身进行挖掘预测的研究，并没有考虑降雨量序列复杂的时频成分及时频差异，蕴含在数据内部的关键信息并不能完全被挖掘，因而降雨预测性能不突出。

小波分析是对信号时间、尺度分析的一种有效方法，其窗口面积不变而时间、频率窗可变[9]，具有多分辨率分析的特性，可以逐步细微的观察、提取数据的特征，使得其对信号具有良好的自适应性，可以很好的解决蕴含在数据中的非平稳时频局域性质难以表达的问题，并且可以容易的把有效数据和噪声分开，根据有效数据与噪声在小波分解时所表现出的不同特征进行有效去噪，从而获得好的去噪效果[10-13]。

因此，本文提出将多细胞基因表达式编程 MC_GEP 算法与小波分析结合进行降水预测建模，很好的解决了直接用 GEP 进行降水数据挖掘预测效果欠佳的问题，并用三组真实的数据实验验证了算法的有效性。

## 2 相关理论基础
### 2.1 小波分解与重构

小波构造的主流方法是多分频率分析，通过塔式多分频率分析 Mallat 算法实现，其核心思想是将信号分解成细节信号和逼近信号的形式，具体分解过程如图 1 所示：

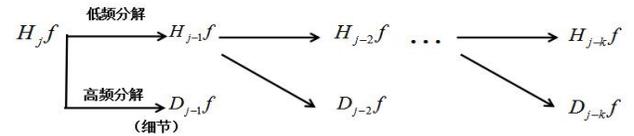

Fig.1 Data signal different frequency decomposition diagram.
图 1 数据信号不同频率分解图

其中能量有限信号 $f \in L^2(R)$ 在分辨率 $2^j$ 下的近似为 $H_j f$，$H_j f$ 可以分解成 $H_{j-1} f$ 与 $D_{j-1} f$ 的和的形式，$H_{j-1} f$ 是 $H_j f$ 通过低通滤波器得到的逼近信号，$D_{j-1} f$ 是 $H_j f$ 通过高通滤波器得到的细节信号，再分解时只将逼近信号分解成细节信号和逼近信号，直到满足分解要求。

Mallat 算法分解过程可以式（2-1）所示：

$$\begin{cases} H_{j+1,k} = \sum_m h_0(m-2k) H_{j,m} \\ D_{j+1,k} = \sum_m h_1(m-2k) H_{j,m} \end{cases} \quad (2-1)$$

其中，$H_j$，$D_j$ 是小波系数的列向量形式；$h_0$ 是低通滤波器的冲击响序列，$h_1$ 是高通滤波器的冲击相应序列。令 $h_0(k) = h_k$，$h_1(k) = g$，则 $g_k = (-1)^k h_{-k+1}$

重构是对分解的逆过程，重构公式为式（2-2）：

$$C_{j-1,k} = \sum_m h_0(k-2m) C_{j,m} + \sum_m h_1(k-2m) D_{j,m} \quad (2-2)$$

重构过程如图 2 所示：

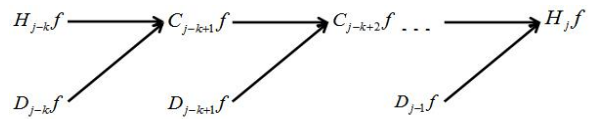

Fig.2 Wavelet reconstruction process
图 2 小波重构过程

## 2.2 多细胞基因表达式编程

多细胞基因表达式编程（Multicellular Gene Expression Programming，MC_GEP）是在 GEP 的基础上引入同源基因和细胞系统概念的改进 GEP 算法。它的编码形式相较于 GEP 更富有弹性，优秀的结构得以延续，搜索空间得以扩大，解的多样性得以增加。

MC_GEP 由多个普通基因和同源基因构成复杂的个体的改进 GEP 算法[14]。同源基因的终结符集和运算符集中每个字符分别表示一种基因和连接基因的符号，使得不同简单基因相结合组成一个复杂的染色体。这种多个基因构成的复杂个体的大大简化了构建一个功能强大的基因型/表现型系统的过程。如图 3 所示的是一个多细胞 GEP 染色体基因型由两个普通基因和一个同源基因组合构成的（为便于理解，图 3 中第一行的数字仅用于标出各基因的基因位，第 2 行是染色体的基因型编码串，并用 2 条虚线将各基因分开）。其中普通基因头长为 8，同源基因头长为 4，普通基因和同源基因的终结符集 T 分别为 $\{?,a,b,c,d,e,f\}$ 和 $\{0,1\}$，其中 '0' 代表第一个普通基因，'1' 代表第二个普通基因，普通基因和同源基因的运算符集 F 均为 $\{+,-,*,/,S,C,q\}$。DC 域为 $\{A,B,C,D,E,F,G,H,I,J\}$，每个字符其对应的常数随机产生。该染色体解码所得表达式树（Expression Tree, ET）为图 4 所示，最终解码的数学表达式为公式（2-3）。

```
0123456789012345678901234561012345678901234567890123450123456789
+-*cda?babafdecbdGGDFDBDCG|/Sb+?debfdfabcddaBHCEDIBCF|+*10111110
```

Fig.3　Multicellular chromosome coding structure

图 3　多细胞染色体编码结构

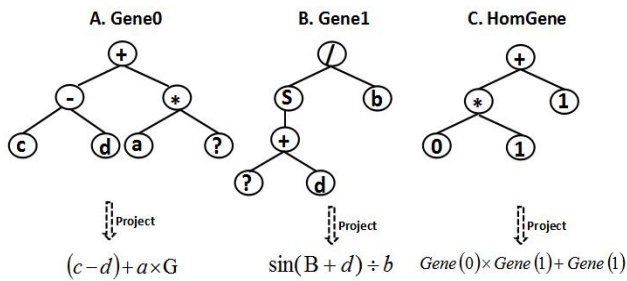

Fig.4　Expression tree of multicellular chromosomes

图 4　多细胞染色体的表达式树

$$((c-d)+a\times G+1)\times(\sin(B+d)\div b) \quad (2-3)$$

## 3 基于 MC_GEP 与小波分析的降水预测算法（WTGEPRP）

降雨量受多种复杂因素共同影响，不断变化的波动幅度及波动频率致使降水时间序列数据呈现非线性、非平稳性的特点。用 GEP 直接对降水时间序列进行分析时数据的细节变化特征难以分析出来，预测的降水量难以精确。且受采集员的经验认知、设备等因素的影响，数据在采集的过程中存在一些噪声是在所难免的，但这些噪声隐含在数据中会直接对数据拟合造成干扰，甚至会影响 GEP 进行函数挖掘，影响预测效果。因此，需要用能处理非线性、非平稳信号并且能有效消除噪声的小波分析方法对数据进行分解和重构，以降低直接对降雨数据建模预测产生的偏差。

### 3.1 WTGEPRP 基本思想

WTGEPRP 算法的基本思想是：首先，降水时间序列根据选定的小波基和分解层数进行小波分解，得到相应的高频分量 $D_1, D_2, \cdots D_n$ 和低频分量 $A_n$。然后，对各高频、低频分量的训练样本分别滑动窗口建模后进行 MC_GEP 算法演化，演化出各分量训练样本的拟合表达式。最后，根据各分量的拟合表达式求解出各分量的预测结果，并对各分量的预测结果进行小波重构得出测试样本结果，即预测数据。具体如图 5 所示：

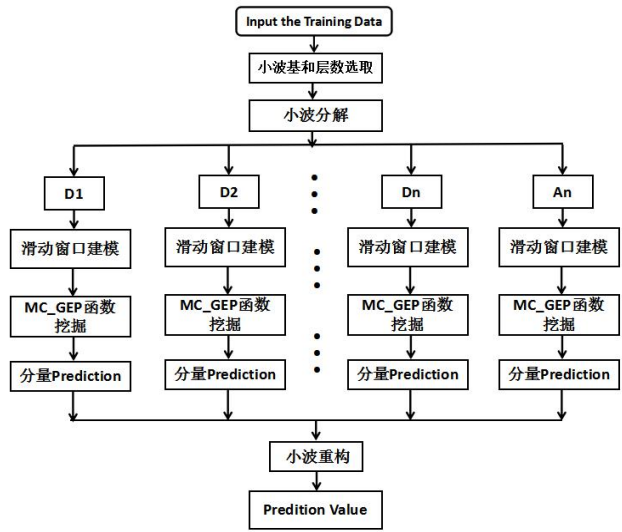

Fig.5 Basic idea of the algorithm

图 5 WTGEPRP 算法基本思想示意图

## 3.2 数据预测建模
### 3.2.1 降水时间序列小波分解

降水时间序列因非线性、非平稳性造成数据特征难以提取，经过小波分解即可容易分析其随机性、周期性和趋势性，并且可以根据降水时间序列中的有效数据与噪声在分解时所表现出的不同特征可以轻易将有效数据和噪声分开进行去噪。

小波分解的关键在于小波基和分解层数的选取，本文通过 MATLAB 小波工具箱和给定的小波基和分解层数对每个实验数据进行小波分解，分解出频率更小的高频分量 $D_1, D_2, \cdots D_n$ 和低频分量 $A_n$（其中 n 为分解层数），以便于更细致的观察降水数据的特征。

### 3.2.2 滑动窗口建模

滑动窗口预测法相比于传统统计学方法更为简单，易理解，无须依赖先验知识和提前知道大致的数学模型，特别适用于 GEP 这样不需要提前假设数据满足的模型的算法，因此本文采用滑动窗口建模预测方法对降雨时间序列进行建模。

设 $X(t)$ 是降雨时间序列，t 是时间（本文是"年"），$x_t$ 是 t 时间对应的降水量，如果滑动窗口大小为 $t-n+1$，就是由降雨数据过去 $t-n$ 个时刻的值预测当前 $t-n+1$ 时刻的值，则建模后时间序列中各数据之间的关系可建立矩阵如式（3-1）所示，目标函数记为式为（3-2）：

$$X(t) = \begin{vmatrix} x_1 & x_2 & \cdots & x_{t-n+1} \\ x_2 & x_3 & \cdots & x_{t-n+2} \\ \cdots & \cdots & \cdots & \cdots \\ x_n & x_{n+1} & \cdots & x_t \end{vmatrix} \quad (3-1)$$

$$x_{t-n+1} = f(x_1, x_2, \cdots, x_{t-n}) \quad (3-2)$$

本文将小波分解后得到的高频分量 $D_1, D_2, \cdots D_n$ 和低频分量 $A_n$ 的训练样本分别通过滑动窗口建模，窗口大小设置为 7，即前六个数进预测当前数据，公式（3-2）中的 $t-n$ 为 6，则矩阵为（3-3）所示，目标函数由式（3-4）表示：

$$X(t) = \begin{vmatrix} x_1 & x_2 & x_3 & x_4 & x_5 & x_6 & x_7 \\ x_2 & x_3 & x_4 & x_5 & x_6 & x_7 & x_8 \\ \cdots & \cdots & \cdots & \cdots & \cdots & \cdots & \cdots \\ x_n & x_{n+1} & \cdots & \cdots & \cdots & \cdots & x_t \end{vmatrix} \quad (3-3)$$

$$x_t = f(x_{t-6}, x_{t-5}, \cdots, x_{t-1}) \quad (3-4)$$

### 3.2.3 MC_GEP 模型挖掘

通过 GEP 对降雨时间序列预测实质上是用 GEP 做函数挖掘，挖掘出具有最好适应度的降水拟合表达式，即挖掘公式（3-4）的具体表达形式，然后以

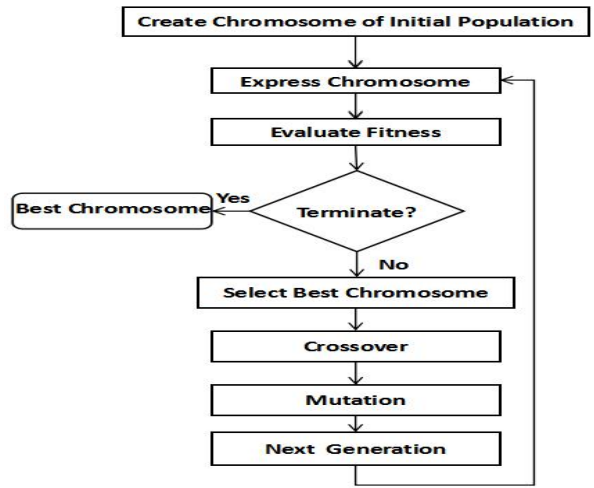

Fig.6 MC_GEP algorithm function mining flow chart

图 6 MC_GEP 算法函数挖掘流程图

此函数模型进行降水值计算与预测。本文通过MC_GEP算法分别对小波分析所得的各分量训练样本进行函数挖掘，从而得到各分量对应的子模型 $f_{D_1}, f_{D_2}, \cdots, f_{D_n}, f_{A_n}$。MC_GEP算法函数挖掘流程如图6所示：

### 3.2.4 小波重构求解预测值

结合MC_GEP函数挖掘出的各分量训练样本拟合表达式和训练样本数据计算出各分量测试样本相对应年份的值。对各分量的预测样本相对应年份的值依据同年份相加的原则进行小波重构，可得到实验数据测试样本的最终相应年份的预测值值。

### 3.3 WTGEPRP算法过程

具体的WTGEPRP算法描述如下：

```
算法    WTGEPRP降水建模预测算法
输入：降水数据，种群大小N, T, F, (同源)基因头长，交叉、变异率
输出：Prediction Value
BEGIN
Step 1  Wavelet analysis
        //小波分解的每个分量的训练集MC_GEP函数挖掘
Step 2    Sliding Window
Step 3    Initialize the GEP Population    //随机产生初始种群
Step 4    while (The termination conditions are not met)
Step 5        Convert the Chromosomes    //解析染色体
Step 6        Evaluate the Fitness    //适应度评价
Step 7        if(Meet termination conditions)
Step 8            return 拟合函数
              //返回最优染色体，即拟合函数
Step 9        else
Step 10           Keep the BestChromosome    //保存上一代最优个体
Step 11           Apply Genetic operations on Population    //种群执行遗传操作
Step 12       end if
Step 13   end while
          //求当前分量的预测集的值
Step 14   Find the value of the prediction set of the current component
          //小波重构
Step 14   The values of all the component prediction sets are added to the year
END
```

## 4 实验及结果分析

### 4.1 实验数据和相关参数

本文通过年郑州1951~2014年年降水量、南宁1951~2013年6月分降水量和澳大利亚墨尔本1911~2012年冬季月均降水量三组经纬度不同和气候差异大的降水量时间序列进行实验，验证算法的效果。具体实验数据集如图7所示，数据集的统计参数如表4-1所示。

Table4-1 Statistical parameters of three sets of data
4-1 数据集的统计参数

| 数据 | 参数简称 | 长度 | 特征 | 最大值(mm) | 最小值(mm) | 均值(mm) | 极值比 |
|---|---|---|---|---|---|---|---|
| 郑州 | Zz | 64 | 年降水量 | 1043 | 353 | 634.828 | 2.955 |
| 南宁 | Nn | 63 | 6月分降水量 | 508 | 30 | 223.381 | 16.933 |
| 墨尔本 | Mb | 102 | 冬季月均降水量 | 341 | | 48.326 | 8.391 |

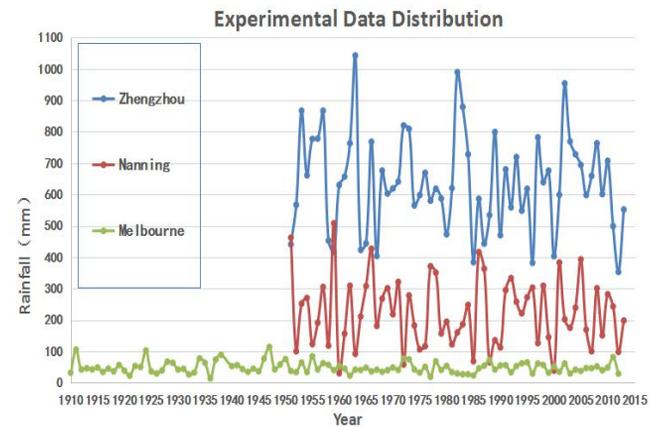

Fig.7    Experimental data distribution
图7  实验数据分布图

分别用GEP、MC_GEP、BP、SVM、EMGEP2PR[15]和WTGEPRP算法以每组数据集的85%为实验的训练样本进行试验，剩余15%为测试样本与预测值进行对比，评比各算法的预测效果。其中GEP相关算法主要参数如表4-2所示，其中终结符 $a, b, c, d, e, f$ 分别与滑动窗口模型中的数据 $x_k$ 的前六个数据 $x_{k-6}$，$x_{k-5}$，$x_{k-4}$，$x_{k-3}$，$x_{k-2}$，$x_{k-1}$ 对应。BP模型调用MATLAB的BP工具箱实现[16]。SVM算法因灵活性和实用性而日渐在降水预测领域运用，并且有比神经网络更小的预测误差[17,18]，本文根据现阶段SVM算法在降水预测中的研究现状，参考其中的参数组合[18~20]，通过软件包LIBSVM-3.22实现，其中核函数设置类型和SVM设置类型采用默认值。并且本文各

算法都以 $\frac{1}{1+RMSE_i}$ 为适应度函数。

Table 4-2 The parameters of GEP correlation algorithm
表 4-2 GEP 相关算法的参数

| Parameter \ Algorithm | GEP | MC_GEP | EMGEP2PR |
|---|---|---|---|
| Gene Head Length | 8 | 8 | 8 |
| Home Gene Head Length | NULL | 4 | 4 |
| Terminal Set | {a,b,c,d,e,f} | {?,a,b,c,d,e,f} | {?,a,b,c,d,e,f} |
| DC Mutation Rate | NULL | 0.25 | 0.25 |
| Gene Mutation Rate | 0.3 | 0.3 | 0.3 |
| Home Gene Mutation Rate | NULL | 0.2 | 0.2 |
| Crossover Rate | 0.3 | | |
| Constant Mutation | 0.05 | | |
| Function Set | {+,-,*,/,sin,cos,sqrt} | | |
| Constant Array Length | 10 | | |
| Population Size | 100 | | |
| Select Mode | Championships Select | | |
| Iteration | 2000 | | |
| Fitness Function | $\frac{1}{1+RMSE}$ | | |

## 4.2 实验结果分析

WTGEPRP 算法进行降雨预测的关键在于选择合适的小波基和分解层数，但是目前并没有统一有效的选取规则。因此，只能通过实验结果与理论结果的误差对比来决定具体的选取。正交小波基分解出的各成分两两正交，不存在线性相关关系，信息没有冗余，能更进一步的分析出数据各层次的细微变化，更好的了解信号变化情况，并能对提取出的各个层次的系数进行重构[21]，因此本文选取 haar、db10、sym8、coif5 四个正交小波基，分解 5 层，将三组数据分别进行实验，并根据实验结果发现三组数据在 coif5 小波基分解 4 层时效果好，因此将选取 coif5 小波基分解 4 层的 RMSE、MAE 作为 WTGEPRP 算法的预测效果，并分别与 GEP、MC_GEP、BP、SVM 和当前 state-of-the-art 的 EMGEP2PR 算法对比，如表 4-3 所示。其中 SVM 的结果值是选取分别参照文献[18]、[19]和[20]中的各参数组合进行实验所得结果中最优值）。

Table 4-3 The experimental results of the three sets of data were compared
表 4-3 各算法对三组数据建模拟合预测实验结果对比表

| 算法 评价指标 | | GEP 拟合 | GEP 预测 | BP[16] 拟合 | BP[16] 预测 | MC_GEP 拟合 | MC_GEP 预测 | SVM[19] 拟合 | SVM[19] 预测 | EMGEP2PR[15] 拟合 | EMGEP2PR[15] 预测 | WTGEPRP 拟合 | WTGEPRP 预测 |
|---|---|---|---|---|---|---|---|---|---|---|---|---|---|
| RMSE | Zz | 176.651 | 186.250 | 158.677 | 168.364 | 149.009 | 156.271 | 131.719 | 138.452 | 89.431 | 92.096 | 52.365 | 55.974 |
| RMSE | Nn | 106.780 | 118.287 | 92.205 | 102.251 | 85.800 | 97.933 | 81.605 | 91.327 | 64.952 | 71.365 | 44.451 | 53.577 |
| RMSE | Mb | 20.298 | 25.744 | 18.444 | 23.783 | 17.342 | 21.229 | 15.377 | 19.143 | 12.602 | 15.114 | 8.539 | 11.991 |
| MAE | Zz | 146.998 | 158.954 | 126.578 | 134.852 | 116.145 | 122.374 | 89.202 | 94.833 | 74.434 | 79.857 | 41.227 | 44.519 |
| MAE | Nn | 88.177 | 99.712 | 83.786 | 94.691 | 75.628 | 84.271 | 72.137 | 80.55 | 53.346 | 59.289 | 34.333 | 43.006 |
| MAE | Mb | 16.064 | 17.381 | 14.156 | 16.342 | 12.960 | 13.537 | 9.261 | 10.93 | 8.546 | 9.035 | 6.431 | 6.944 |

从表中的数据得出：

（1）依据评价指标 RMSE 或 MAE 可知 WTGEPRP 算法拟合和预测效果最好。

（2）对比 WTGEPRP 算法与其他算法对三组不同降水数据集上进行建模预测的 RMSE 值：

A、郑州降水数据集的实验，WTGEPRP 模型预测获得到的 RMSE 值分别比 GEP、BP、MC_GEP、SVM、EMGEP2RP 预测获得 RMSE 的降低了 71.993%、66.987%、63.621%、59.572%、44.252%；

B、南宁降水数据集的实验，WTGEPRP 模型预测获得到的 RMSE 值分别比 GEP、BP、MC_GEP、SVM、EMGEP2RP 预测获得 RMSE 的降低了 54.706%、47.602%、45.292%、41.335%、24.925%；

C、墨尔本降水数据集的实验，WTGEPRP 模型预测获得到的 RMSE 值分别比 GEP、BP、MC_GEP、SVM、

EMGEP2RP 预测获得 RMSE 的降低了 53.4221%、49.582%、43.516%、37.361%、20.663%。

（3）对比 WTGEPRP 算法与其他算法对三组不同降水数据集上进行建模预测的 MAE 值：

A、郑州降水数据集的实验，WTGEPRP 模型预测获得到的 MAE 值分别比 GEP、BP、MC_GEP、SVM、EMGEP2RP 预测获得 MAE 的降低了 71.993%、66.987%、63.621%、53.055%、44.252%；

B、南宁降水数据集的实验，WTGEPRP 模型预测获得到的 MAE 值分别比 GEP、BP、MC_GEP、SVM、EMGEP2RP 预测获得 MAE 的降低了 56.87%、54.583%、48.967%、46.61%、27.464%；

C、墨尔本降水数据集的实验，WTGEPRP 模型预测获得到的 MAE 值分别比 GEP、BP、MC_GEP、SVM、EMGEP2RP 预测获得 MAE 的降低了 60.048%、57.508%、48.704%、36.468%、23.143%。

（3）、数据极值比越大 WTGEPRP 算法预测的 RMSE、MAE 值比其他算法降低的越大，预测误差越小，效果越好。

## 5 结束语

本文提出一种基于小波分析和多细胞基因表达式编程的降水预测算法 WTGEPRP，并将其与 GEP 算法、BP 算法、MC_GEP 算法、SVM 算法和 EMGEP2RP 算法通过郑州、南宁、澳大利亚墨尔本历史降水数据进行模拟实验效果分析对比，WTGEPRP 算法的 RMSE、MAE 评价指标都是最小，拟合与预测效果都好于其它算法，充分证明了 WTGEPRP 算法的有效性，说明了通过 MC_GEP 与小波分析结合的方法确实能挖掘数据深层隐含的信息，弥补了 GEP 算法直接对降雨量时间序列预测时忽视序列复杂的时频成分和时频差异的缺陷。虽然小波分析中小波函数的选择需要通过实验不断尝试，致使 WTGEPRP 算法运行时间长，计算量大，但气候降水预测的实际应用中主要强调预测的精度而时间效率几乎可忽略，可见，WTGEPRP 算法将具有较好的气象应用前景，而且也可以迁移到其它时间序列预测问题中。